\documentclass[runningheads]{llncs}
\usepackage{graphicx}
\begin{document}
\title{Recent Advances in Video Question Answering: A Review of Datasets and Methods}
\titlerunning{Recent Advances in Video Question Answering}
\author{Devshree Patel \and
Ratnam Parikh \and
Yesha Shastri} 
\authorrunning{D. Patel et al.}
\institute{School of Engineering and Applied Science, Ahmedabad University, India}
\maketitle              % typeset the header of the contribution
\begin{abstract}
Video Question Answering (VQA) is a recent emerging challenging task in the field of Computer Vision. Several visual information retrieval techniques like Video Captioning/Description and Video-guided Machine Translation have preceded the task of VQA. VQA helps to retrieve temporal and spatial information from the video scenes and interpret it. In this survey, we review a number of methods and datasets for the task of VQA. To the best of our knowledge, no previous survey has been conducted for the VQA task.
\end{abstract}
\section{Introduction}
Computer Vision is one of the sub-fields of Artificial Intelligence. It enables computers to interpret and analyze real-world scenes as humans do. Some of the computer vision research tasks are segmentation, object detection/ tracking, visual captioning and VQA. Out of them, VQA is a more challenging task since it requires an understanding of visual data based on the question asked in natural language. Since, most of the visual information we face in the real world is either in the form of images or videos, the task of VQA will help to extract crucial information for building real-life applications.

The task of VQA can be bifurcated into two sub-tasks: multiple-choice QA and open-ended QA. The latter is more challenging since it does not have a fixed answer set. Also, an in-depth understanding of the video scene is required to produce an answer from the pool of all possible answers. Open-ended QA finds its applicability in the real-world since it helps to understand complex scene and generate relevant information without the constraints of choices.

The collection of data for image-based QA is relatively easier as compared to video-based QA. The creation of datasets for VQA is a burdensome task as it requires huge human efforts for creating QA pairs and accurate annotations by looking at the visual content. Additionally, video content must have varied actions, scenes and object interactions in order to build a robust VQA model. In general, it is a difficult task to derive interpretations from videos since they comprise of varied length having multiple frames. 

The models for VQA have been created by extending the models for image question answering. Earlier approaches for constructing models of VQA only considered the spatial information from the videos. However, VQA requires to find clues accurately on both spatial and temporal dimensions. Proceeding with the methods, [3] introduced the approach for extracting temporal context information from the videos. Further, [5] studied the problem of modeling temporal dynamics with frame-level attention by proposing an attribute augmented attention network for VQA. In [7], authors exploit information relating to appearance and motion residing in the video with an optimized attention mechanism. In contrast to the aforementioned approaches, [16] presents a novel approach of positional self-attention by replacing RNNs with self-attention. [12] proposes an architecture based on Multimodal Dual Attention Memory (MDAM) to extract the latent concepts from video scenes and captions.Unlike the previous approaches which apply spatio-temporal attention, [21] improves the approach by introducing location and relations among object interactions by creating a location-aware graph network. In [27] authors leverage the task of VQA by using Bidirectional Encoder Representations (BERT) for capturing the attention of every word from both the directions.

\section{Datasets}
Data acquisition for VQA has been a challenging task due to requirement of large video corpus with varied content. One of the first datasets for VQA is YouTube2Text [48] which comprises of 1987 videos and 122,708 descriptions of those in natural language. Many existing models [5, 9, 13, 21] address the task of VQA with the help of YouTube2Text [48]. The LSMDC 2016 description dataset [47] enables multi-sentence description of the videos which is a unique property from the previous approaches of datasets. It comprises of $\sim 128$K clips taken as a combination from M-VAD and MPII-MD datasets. Based on the LSMDC 2016 description dataset [47], a fill-in-the-blank dataset (MovieFIB) [42] was proposed. This was a simple benchmark dataset consisting of 300,000 fill-in-the-blank question-answer and video pairs. To answer these fill-in-the-blank questions a model is expected to have basic understanding of scenes by detecting activities, objects, people and their interactions. Table 5 shows dataset statistics for existing VQA datasets.

\subsection{MovieQA}
MovieQA [36] is one of the unique benchmark datasets for VQA as it contains multiple sources of information. It comprises of video clips, subtitles, plots and scripts. It is aimed to perform automatic story comprehension from both text and video. The dataset includes 408 movies and 14,944 questions with semantic variations. [8,12,20,23] attempt to tackle the VQA task with MovieQA dataset [36]. The original paper [36] obtains an accuracy of $\sim35\%$ which is gradually improved and reached to $\sim45\%$ by [23]. A comparison of various models using MovieQA dataset is shown in Table 1.
\subsection{YouTube2TextQA}
YouTube2TextQA was one of the first dataset VQA datasets proposed by [48]. The data is taken from the YouTube2Text dataset. It comprises of 1987 videos and 122,708 automatically generated QA pairs. The dataset can address multiple-choice questions. Initially, an accuracy of 61.4\% was obtained with r-ANL [5] method and it reached to 83.9\% with one of the latest models L-GCN [21].
\begin{table}[!ht]
     \caption{Best Test Accuracies of models using MovieQA and Youtube2TextQA datasets}
     \makebox[\linewidth]{
     \begin{tabular}{|c|c|c|c|c|c|c|c|c|}
\hline
\textbf{Datasets}          & \multicolumn{4}{c|}{\textit{\textbf{YouTube2TextQA}}}                                   & \multicolumn{4}{c|}{\textit{\textbf{MovieQA}}}                                        \\ \hline
\textit{\textbf{Model}}    & r-ANL {[}5{]} & DLAN {[}9{]} & Fan et al. {[}13{]} & L-GCN {[}21{]} & DEMN {[}8{]} & MDAM {[}12{]} & PAMN {[}20{]} & AMN {[}23{]} \\ \hline
\textit{\textbf{Accuracy}} & 61.4              & 36.33               & 82.5                & 83.9                  & 30                 & 41.41               & 42.53               & 45.31                \\ \hline
\end{tabular}
     }
\end{table}
\subsection{MSRVTT-QA and MSVD-QA}
MSRVTT-QA and MSVD-QA were first used by [7] based on MSRVTT [45] and MSVD [49] video datasets. MSRVTT-QA consists of $\sim 10$K video clips and $\sim 243$K QA pairs. MSVD-QA is mainly used in video captioning experiments but due to its large data size it is also used for VQA. It has 1970 video clips and $\sim 50.5$K QA pairs. A comparison of various models using MSRVTT-QA and MSVD-QA datasets is shown in Table 2. 
 \begin{table}[!ht]
     \caption{Accuracies of models using MSRVTT-QA(Left) and MSVD-QA(Right)}
     \makebox[\linewidth]{
   \begin{tabular}{|l|l|l|l|l|l|l|l|l|l|l|l|l|}
\hline
                       & \multicolumn{6}{l|}{MSRVTT-QA}           & \multicolumn{6}{l|}{MSVD-QA}             \\ \hline
Model                  & What & Who  & How  & When & Where & All  & What & Who  & How  & When & Where & All  \\ \hline
Xu et al. {[}7{]}      & 26.2 & 43   & 80.2 & 72.5 & 30    & 32.5 & 20.6 & 47.5 & 83.5 & 72.4 & 53.6  & 32   \\ \hline
Fan et al. {[}13{]}    & 26.5 & 43.6 & 82.4 & 76   & 28.6  & 33   & 22.4 & 50.1 & 73   & 70.7 & 42.9  & 33.7 \\ \hline
CAN {[}22{]}           & 26.7 & 43.7 & 83.7 & 75.3 & 35.2  & 33.2 & 21.1 & 47.9 & 84.1 & 74.1 & 57.1  & 32.4 \\ \hline
Resnet+CDC(AA){[}15{]} & 27.3 & 44.2 & 86.3 & 73.9 & 34.4  & 34.1 & -    & -    & -    & -    & -     & -    \\ \hline
QueST {[}28{]}         & 27.9 & 45.6 & 83   & 75.7 & 31.6  & 34.6 & 24.5 & 52.9 & 79.1 & 72.4 & 50    & 36.1 \\ \hline
TSN {[}17{]}           & 27.9 & 46.1 & 84.1 & 77.8 & 37.6  & 35.4 & 25   & 51.3 & 83.8 & 78.4 & 59.1  & 36.7 \\ \hline
Jin et al. {[}19{]}    & 29.5 & 45   & 83.2 & 74.7 & 42.4  & 35.4 & 24.2 & 49.5 & 83.8 & 74.1 & 53.6  & 35   \\ \hline
Jiang et al. {[}31{]}  & 29.2 & 45.7 & 83.5 & 75.2 & 34    & 35.5 & 23.5 & 50.4 & 83   & 72.4 & 46.4  & 34.7 \\ \hline
HCRN {[}29{]}          & -    & -    & -    & -    & -     & 35.6 & -    & -    & -    & -    & -     & 36.1 \\ \hline
MHMAN {[}30{]}         & 28.7 & 47.1 & 85.1 & 77.1 & 35.2  & 35.6 & 23.3 & 50.7 & 84.1 & 72.4 & 53.6  & 34.6 \\ \hline
Zhang et al. {[}15{]}  & -    & -    & -    & -    & -     & -    & 21.3 & 48.3 & 82.4 & 70.7 & 53.6  & 32.6 \\ \hline
\end{tabular}
}
\end{table}
\subsection{VideoQA}
VideoQA dataset [4] consists of the questions in free-form natural language as opposed to previous fill-in-the-blank questions [42]. The dataset is made from the internet videos having user-curated descriptions. QA pairs are generated automatically from the descriptions by using a state-of-the-art question generation method. A collection of 18,100 videos and 175,076 candidate QA pairs make up this large-scale dataset for VQA. 
\subsection{Pororo-QA}
In [8], authors construct a large-scale VQA dataset - PororoQA from cartoon video series. The dataset is made up of videos having simple but a coherent story structure as compared to videos in previous datasets. The dataset includes $\sim 16$K dialogue-scene pairs, $\sim 27$K descriptive sentences and $\sim 9$K multiple-choice questions. DEMN [8] achieved best accuracy of 68\% at the time of release. A comparison of various models using PororoQA dataset is shown in Table 3.
\subsection{TVQA+ and TVQA}
TVQA+ dataset [41] is built upon one of the largest VQA datasets - TVQA [40]. TVQA [40] is a dataset built on natural video content, collected from 6 famous TV series. The videos have rich dynamic interactions and the QA pairs are formed by the people considering both the videos and the accompanying dialogues. The provision of temporal annotations by the TVQA dataset [40] is one of the key properties. However, this dataset does not consider the spatial annotations which are equally important as temporal. Hence, TVQA+ [41] is a spatio-temporally grounded VQA dataset made by adding bounding boxes to the essential regions of the videos from a subset of the TVQA dataset [40]. It comprises of $\sim29.4$K multiple-choice questions annotated in both spatial and temporal domains. A comparison of various models using TVQA dataset is shown in Table 3.
\subsection{ActivityNet-QA}
[35] introduces ActivityNet-QA, a fully manually annotated large-scale dataset. It consists of $\sim58$K QA pairs based on $\sim 5.8$K web videos taken from the popular ActivityNet dataset [50]. The dataset consists of open-ended questions. A comparison of various models using ActivityNet-QA dataset is shown in Table 3.
 \begin{table}[!ht]
     \caption{Best Test Accuracies of models using PororoQA, TVQA and ActivityNetQA datasets}
     \makebox[\linewidth]{
     \begin{tabular}{|c|c|c|c|c|c|c|c|c|}
\hline
\textbf{Datasets}          & \multicolumn{3}{c|}{\textit{\textbf{PororoQA}}}                 & \multicolumn{3}{c|}{\textit{\textbf{TVQA}}}                       & \multicolumn{2}{c|}{\textit{\textbf{ActivityNetQA}}} \\ \hline
\textit{\textbf{Model}}    & DEMN {[}8{]} & MDAM {[}12{]} & Yang et al. {[}27{]} & PAMN {[}20{]} & AMN {[}23{]} & Yang et al. {[}27{]} & CAN {[}22{]}    & MAR-VQA(a+m) {[}25{]}    \\ \hline
\textit{\textbf{Accuracy}} & 68.0               & 48.9               & 53.79                 & 66.77                  & 70.70                & 73.57                & 35.4                  & 34.6                         \\ \hline
\end{tabular}
     }
   \end{table}
\subsection{TGIF-QA}
[44] proposes a TGIF-QA dataset from the earlier (Tumblr GIF) TGIF dataset which was used for video captioning task. TGIF-QA dataset [44] contains $\sim 165$K QA pairs from $\sim 72$K animated GIFs taken from the TGIF dataset. The TGIF dataset was ideal for VQA task as it comprised of attractive animated GIFs in a concise format for the purpose of story telling. A comparison of various models using TGIF-QA dataset is shown in Table 4 (Lower the count value, better is the model performance).
 \begin{table}[!ht]
     \caption{Models using TGIF-QA dataset}
     \makebox[\linewidth]{
     \begin{tabular}{|c|c|c|c|c|}
\hline
Model    & \multicolumn{1}{l|}{Action} & \multicolumn{1}{l|}{Trans} & \multicolumn{1}{l|}{Frame} & \multicolumn{1}{l|}{Count} \\ \hline
MAR-VQA(a+m) {[}25{]} & 71.1                        & 71.9                       & 56.6                       & 5.35                       \\ \hline
PSAC {[}16{]} & 70.4                        & 76.9                       & 55.7                       & 4.27                       \\ \hline
STA(R) {[}18{]} & 72.3                        & 79                         & 56.6                       & 4.25  \\ \hline
Jin et al. {[}19{]} & 72.7                        & 80.9                       & 57.1                       & 4.17                       \\ \hline
Co-Mem {[}10{]} & 68.2                        & 74.3                       & 51.5  & 4.1   \\ \hline
Jiang et al. {[}31{]} & 75.4                        & 81                         & 55.1                       & 4.06                       \\ \hline
Fan et al. {[}13{]} & 73.9                        & 77.8  & 53.8                       & 4.02                       \\ \hline
L-GCN {[}21{]} & 74.3                        & 81.1  & 56.3                       & 3.95                       \\ \hline
HCRN {[}29{]} & 75                          & 81.4                       & 55.9                       & 3.82                       \\ \hline
\end{tabular}
     }
   \end{table}
\subsection{LifeQA}
The LifeQA dataset from [33] considers the real life scenarios in its video clips unlike the movies or TV shows which contain the edited content. In real-life QA systems, this dataset will be useful because of its relevance to day-to-day lives. It comprises of 275 videos and 2328 multiple-choice questions.
\subsection{DramaQA}
[43] proposes DramaQA dataset which focuses upon character-centred representations with richly annotated 217,308 images taken from the TV drama "Another Miss Oh". The annotations consider the aspects of behaviour and emotions of the characters. In total, the dataset consists of $\sim23$K video clips and $\sim16$K QA pairs.
\subsection{Social-IQ}
Social-IQ dataset [38] also referred as Social Intelligence Queries dataset targets to analyze the unscripted social situations which comprise of natural interactions. The dataset is prepared extensively from the YouTube videos as they are rich in dynamic social situations. This dataset comprising of 1,250 social videos, 7,500 questions and a total of 52,500 answers including 30,000 correct and 22,500 incorrect answers is created with an aim to benchmark social intelligence in AI systems.
\subsection{MarioQA}
The main motivation behind creating MarioQA dataset [37] was to understand temporal relationships between video scenes to solve VQA problems. Before MarioQA [37], all the datasets required excessive reasoning through single frame inferences. The dataset was synthesized from super mario gameplay videos. It consists of $\sim92$K QA pairs from video sets as large as 13 hours. 
\subsection{EgoVQA}
The EgoVQA dataset proposed by [34] presents a novel perspective of considering first-person videos as most of the previous VQA datasets [MSRVTT-QA [7], MSVD-QA [7], Youtube2TextQA [48]] focus on third-person video datasets. Answering questions about first-person videos is a unique and challenging task because the normal action recognition and localization techniques based on the keywords in questions will not be effective here. To address the same, 600 question-answer pairs with visual contents are taken across 5,000 frames from 16 ﬁrst-person videos.  
\subsection{Tutorial-VQA}
Tutorial-VQA [32] was prepared to address the need for a dataset comprising of non-factoid and multi-step answers. The technique for data acquisition was same as MovieQA [36] and VideoQA [4] as it was build from movie-scripts and news transcripts. The answers of MovieQA [36] had a shorter span than the answers collected in Tutorial-VQA [32]. In VideoQA dataset [4], questions are based on single entity in contrast to instructional nature of Tutorial-VQA [32]. It consists of 76 videos and 6195 QA pairs. 
\subsection{KnowIT-VQA}
KnowIT VQA (knowledge informed temporal VQA) dataset [39] tries to resolve the limited reasoning capabilities of previous datasets by incorporating external knowledge. External knowledge will help reasoning beyond the visual and textual content present in the videos. The collected dataset comprises of videos annotated with knowledge-based questions and answers from the popular TV show - The Big Bang Theory. There are a total of $\sim 12$K video clips and $\sim 24$K QA pairs, making it one of the largest knowledge-based human-generated VQA dataset. 

 \begin{table*}[!ht]
\caption{Dataset statistics for existing video question answering datasets}
\centering
\begin{tabular}{|l|l|l|l|l|l|}
\hline
\textbf{Dataset}        & \textbf{Videos} & \textbf{QA Pairs} & \textbf{QA Type}                                         & \textbf{Source}                                                               & \textbf{QA Tasks}                                                         \\ \hline
TutorialVQA {[}32{]}    & 76                     & 6195              & Manual                                                         & Screencast Tutorials                                                          & Open-ended                                                                \\ \hline
LifeQA {[}33{]}         & 275                    & 2328              & Manual                                                         & Daily life videos                                                             & Multiple-choice                                                           \\ \hline
EgoVQA {[}34{]}         & 520                    & 600               & Manual                                                         & Egocentric video studies                                                      & \begin{tabular}[c]{@{}l@{}}Open-ended and \\ multiple-choice\end{tabular} \\ \hline
ActivityNetQA {[}35{]}  & 5800                   & 58,000             & Manual                                                         & ActivityNet dataset                                                           & Open-ended                                                                \\ \hline
MovieQA {[}36{]}        & 6771                   & 6462              & Manual                                                         & Movies                                                                        & Multiple-choice                                                           \\ \hline
PororoQA {[}8{]}        & 16,066                  & 8913              & Manual                                                         & Cartoon Videos                                                                & Multiple-choice                                                           \\ \hline
MarioQA {[}37{]}        & -                      & 187,757            & Automatic                                                      & Gameplay videos                                                               & Multiple-choice                                                           \\ \hline
TGIF-QA {[}44{]}        & 71,741                  & 165,165            & \begin{tabular}[c]{@{}l@{}}Automatic and\\ Manual\end{tabular} & \begin{tabular}[c]{@{}l@{}}TGIF(internet animated\\ GIFs)\end{tabular}        & \begin{tabular}[c]{@{}l@{}}Open-ended and \\ multiple-choice\end{tabular} \\ \hline
TVQA {[}40{]}           & 21,793                  & 152,545            & Manual                                                         & \begin{tabular}[c]{@{}l@{}}Sitcoms,medical and \\ criminal drama\end{tabular} & Multiple-choice                                                           \\ \hline
TVQA+ {[}41{]}          & 4198                   & 29,383             & Manual                                                         & \begin{tabular}[c]{@{}l@{}}Sitcoms,medical and \\ criminal drama\end{tabular} & Multiple-choice                                                           \\ \hline
Know-IT-VQA {[}39{]}    & 12,087                  & 24,282             & Manual                                                         & Big Bang Theory                                                               & Multiple-choice                                                           \\ \hline
Movie-FIB {[}42{]}      & 118,507                 & 348,998            & Automatic                                                      & LSMDC 2016                                                                    & Open-ended                                                                \\ \hline
VideoQA {[}4{]}         & 18,100                  & 175,076            & Automatic                                                      & Online internet videos                                                        & Open-ended                                                                \\ \hline
DramaQA {[}43{]}        & 23,928                  & 16,191             & \begin{tabular}[c]{@{}l@{}}Automatic and\\ Manual\end{tabular} & TV Drama                                                                      & Multiple-choice                                                           \\ \hline
Social-IQ {[}38{]}      & 1250                   & 7500              & Manual                                                         & YouTube Videos                                                                & Multiple-choice                                                           \\ \hline
MSRVTT-QA {[}7{]}       & 10,000                  & 243,680            & Automatic                                                      & MSRVTT(Web videos)                                                            & Open-ended                                                                \\ \hline
MSVD-QA {[}7{]}         & 1970                   & 50,505             & Automatic                                                      & MSVD(Web videos)                                                              & Open-ended                                                                \\ \hline
Youtube2TextQA {[}48{]} & 1987                   & 122,708            & Automatic                                                      & Youtube2Text                                                                  & Multiple-choice                                                           \\ \hline
\end{tabular}
\end{table*}

\section{Methods}
VQA is a complex task which that combines the domain knowledge of both computer vision and natural language processing. The methods implemented for VQA tasks have been extended from the methods of Image-QA. VQA is more challenging compared to Image-QA because it expects spatial as well as temporal mapping and requires understanding of complex object interactions in the video scenes. One of the first works on VQA was proposed by [1]. They build a query-answering system based on a joint parsing graph from videos as well as texts. As an improvement over the model proposed by [1], [3] proposes a flexible encoder-decoder architecture using RNN, which covers a wide range of temporal information. Leveraging the task of VQA, [4] proposes a model which is capable of generating QA pairs from learned video descriptions. [4] constructs a large-scale VQA dataset with $\sim18$K videos and $\sim175$K candidate QA pairs. In comparison to previous baseline methods of Image-QA, the model proposed in [4] proves to be efficient. 

Following the aforementioned approaches, several methods have evolved in the literature to solve the complex task of VQA. The first sub-section presents the methods based on spatio-temporal approaches. The second sub-section deals with memory-based methods. Methods using attention mechanism are listed in the third sub-section. The fourth sub-section details the multimodal attention based approaches. In addition to these, various novel approaches for VQA are described in the later section. A summary of results from all the described models is abridged in tables 1, 2, 3 and 4. 

\subsection{Spatio-Temporal Methods}
Joint reasoning of spatial and temporal structures of a video is required to accurately tackle the problem of VQA. The spatial structure will give information about in-frame actions and temporal structure will analyze the sequence of actions taking place in the videos. Methods which work on the same motivation are described in detail in the below sections.
\subsubsection{r-STAN (Zhao et al., 2017)}
The authors of [6] propose a hierarchical Spatio-temporal attention encoder-decoder learning framework to address the problem
of open-ended Video-QA. The proposed model jointly learns the representation of
the sequential frames with targeted objects and performs multi-modal reasoning
according to the text and video. [18] infers the long-range temporal structures
from the videos to answer the open-ended questions.
\subsubsection{STA(R) (Gao et al., 2019)}
A Structured Two-stream Attention network (STA) is developed by [18] with the help of a structured segment component and the encoded text features. The structured two-stream attention component focuses on the relevant visual instances from the video by neglecting the unimportant information like background. The fusion of video and question aware representations generate the inferred answer. Evaluation of the model is conducted on the famous TGIF-QA dataset.
\subsubsection{L-GCN (Huang et al., 2020)}
Along with the extraction of spatio-temporal features, taking into account the information about object interactions in the videos helps to infer accurate answers for VQA. Considering this, [21] proposes a novel and one of the first graph model network called Location-Aware Graph Network. In this approach, an object is associated to each node by considering features like appearance and location. Inferences like category and temporal locations are obtained from the graph with the help of graph convolution. The final answer is then generated by merging the outputs of the graph and the question features. The proposed method [21] is evaluated on three datasets - YouTube2Text-QA [48], TGIF-QA [44], and MSVD-QA [7].
\subsubsection{QueST (Jiang et al., 2020)}
Often more significance is given to spatial and temporal features derived from the videos but less to questions, therefore, accounting to this, the authors of [28] propose a novel Question-Guided Spatio-Temporal contextual attention network (QueST) method. In QueST [28], semantic features are divided into spatial and temporal parts, which are then used in the process of constructing contextual attention in spatial and temporal dimensions. With contextual attention, video features can be extracted from both spatio-temporal dimensions. Empirical results are obtained by evaluating the model on TGIF-QA [44], MSVD-QA [7], and MSRVTT-QA [7] datasets. 
\subsection{Memory-based Methods}
To answer questions for VQA, models need to keep track of the past and future frames as the answers would require inferences from multiple frames in time. Below are descriptions of some of the approaches which implemented memory-based models. 
\subsubsection{DEMN (Kim et al., 2017)}
A novel method based on video-story learning is developed by [8] which uses Deep Embedded Memory Network (DEMN) to reconstruct stories from a joint scene dialogue video stream. The model uses a latent embedding space of data. Long short term memory (LSTM) based attention model is used to store video stories for answering the questions based on specific words. The model consists of three modules: video story understanding, story selection and answer selection. Baseline models such as Bag-of-Words (BOW), Word2vec(W2V), LSTM, end to end memory networks, and SSCB are used to compare with the proposed model. In addition to the proposed model, authors also create PororoQA dataset which is the first VQA dataset having a coherent story line throughout the dataset. %The model attain state-of-the-art results for both MovieQA [36] and PororoQA datasets.
\subsubsection{Co-Mem (Gao et al., 2018)}
A novel method of motion appearance co-memory networks is developed by [10] which considers the following two observations. First, motion and appearance when correlated becomes an important feature to be considered for the VQA task. Second, different questions have different requirements of the number of frames to form answer representations. The proposed method uses two stream models that convert the videos into motion and appearance features. The extracted features are then passed to a temporal convolutional and deconvolutional neural network to generate the multi-level contextual facts. These facts are served as input to the co-memory networks which are built upon the concepts of Dynamic Memory Networks (DMN). Co-memory networks aid in joint modeling of the appearance and motion information with the help of a co-memory attention mechanism. The co-memory attention mechanism takes the appearance cues for generating motion attention and motion cues for generating appearance attention. The proposed model is evaluated on the TGIF-QA [44] dataset. %The model outperforms the state-of-the-art methods on the TGIF-QA [44] dataset. Model outperforms a dual LSTM based method containing both spatial and temporal attention [44] and VIS-LSTM, VQA-MCB methods mentioned in [16].
\subsubsection{AHN (Zhao et al., 2018)}
To tackle the problem of long-form VQA, the authors of [11] create an adaptive hierarchical reinforced network (AHN). The encoder-decoder network comprises adaptive encoder network learning and a reinforced decoder network learning. In adaptive encoder learning, the attentional recurrent neural networks segment the video to obtain the semantic long-term dependencies. It also aids the learning of joint representation of the segments and relevant video frames as per the question. The reinforced decoder network generates the open-ended answer by using the segment level LSTM networks. A new large scale long-form VQA dataset, created from the ActivityNet data [50], is also contributed by [11]. For the long-form VQA task, the proposed model is compared with state-of-the-art methods [STAN [6], MM+ [4] and ST-VQA [44]] on which the new model outperforms.
\subsubsection{Fan et al., 2018}
[13] proposes a novel heterogeneous memory component to integrate appearance and motion features and learn spatio-temporal attention simultaneously. A novel end-to-end VQA framework consisting of three components is proposed. The first component includes heterogeneous memory which extracts global contextual information from appearance and motion features. The second component of question memory enables understanding of complex semantics of questions and highlights queried subjects. The third component of multimodal fusion layer performs multi-step reasoning by attending relevant textual and visual clues with self-attention. Finally, an attentional fusion of the multimodal visual and textual representations helps infer the correct answer. [13] outperforms existing state-of-the-art models on the four benchmark datasets - TGIF-QA [44], MSVD-QA [7], MSRVTT-QA [7], and YouTube2Text-QA [48].
\subsubsection{MHMAN (Yu et al., 2020)}
Humans give attention only to certain important information to answer questions from a video rather than remembering the whole content. Working on the same motivation, [30] proposes a framework comprising of two heterogeneous memory sub-networks. The top-guided memory network works at a coarse-grained level which extracts filtered-out information from the questions and videos at a shallow level. The bottom memory networks learn fine-grained attention by guidance from the top-guided memory network that enhances the quality of QA. Proposed MHMAN [30] model is evaluated on three datasets - ActivityNet-QA [35], MSRVTT-QA [7] and MSVD-QA [7]. Results demonstrate that MHMAN [30] outperforms the previous state-of-the-art models such as CoMem [10] and HME [13]. 
\subsection{Attention-based Methods}
Attention mechanism plays an important role in VQA as it helps to extract answers efficiently from the videos by giving more substance to certain relevant words from the questions. The approaches mentioned below develop models by exploiting the different aspects of attention mechanism.
\subsubsection{r-ANL (Ye et al., 2017)}
As an improvement to the earlier presented approach [3] which considered temporal dynamics, [5] also incorporates frame-level attention mechanism to improve the results. [5] proposes an attribute-augmented attention network that enables the joint frame-level attribute detection and unification of video representations. Experiments are conducted with both multiple-choice and open-ended questions on the YouTube2Text dataset [50]. The model is evaluated by extending existing Image-QA methods and it outperforms the previous Image-QA baseline methods.
\subsubsection{Xu et al., 2017}
In [7], authors exploit the appearance and motion information resided in the video with a novel attention mechanism. To develop an efficient VQA model, [7] proposes an end-to-end model which refines attention over the appearance and motion features of the videos using the questions. The weight representations and the contextual information of the videos are used to generate the answer. [7] also creates two datasets - MSVD-QA and MSRVTT-QA from the existing video-captioning datasets - MSVD [49] and MSRVTT [45]. %The model yields approximately 32\% accuracy in all tasks of the MSVD-QA and MSRVTT-QA datasets.
\subsubsection{DLAN (Zao et al., 2017)}
In [9], authors study the problem of VQA from the viewpoint of Hierarchical Dual-Level Attention Network (DLAN) learning. On the basis of frame-level and segment-level feature representations, object appearance and movement information from the video is obtained. The object appearance is obtained from frame-level representations using a 2D-ConvNet whereas movement information based on segment-level features is done using a 3D-ConvNet. Dual-level attention mechanism is used to learn the question-aware video representations with question-level and word-level attention mechanisms. The authors also construct large scale VQA datasets from existing datasets - VideoClip [51] and YouTube2Text [48]. %The model yields an overall accuracy of 36\% on Youtube2Text [48] and 55\% on VideoClip [51] dataset.
\subsubsection{Resnet+CDC (AA) (WenqiaoZhang et al., 2019)}
Inspired from previous contributions based on attention mechanisms in VQA systems, [15] studies the problem related to unidirectional attention mechanism which fails to yield a better mapping between modalities. [15] also works on the limitation of previous works that do not explore high-level semantics at augmented video-frame level. In the proposed model, each frame representation with context information is augmented by a feature extractor- (ResNet+C3D variant). In order to yield better joint representations of video and question, a novel alternating attention mechanism is introduced to attend frame regions and words in question in multi-turns. Furthermore, [15] proposes the first-ever architecture without unidirectional attention mechanism for VQA. %The model outperforms previous state-of-the-art frameworks on MSVD-QA [7] and MSRVTT-QA [7] datasets. In comparison to [7] and [13], [15] yields an increase in accuracy by 1\%.
\subsubsection{PSAC (Li et al., 2019)}
In contrast to previous works like [10], [16] proposes Positional Self Attention with Co-attention (PSAC) architecture without recurrent neural networks for addressing the problem of VQA. Despite the success of various RNN based models, sequential nature of RNNs pose a problem as they are time-consuming and they cannot efficiently model long range dependencies. Thus, positional self attention is used to calculate the response at each position by traversing all positions of sequence and then adding representations of absolute positions. PSAC [16] exploits the global dependencies of temporal information of video and question, executing video and question encoding processes in parallel. In addition to positional attention, co-attention is used to attend relevant textual and visual features to guarantee accurate answer generation. Extensive experimentation on TGIF-QA [44] shows that it outperforms RNN-based models in terms of computation time and performance. 
\subsubsection{TSN (Yang et al., 2019)}
The authors of [17] work on the failure of existing methods that take into account the attentions on motion and appearance features separately. Based on the motivation to process appearance and motion features synchronously, [17] proposes a Question-Aware Tube Switch Network (TSN) for VQA. The model comprises of two modules. First, correspondence between appearance and motion is extracted with the help of a mix module at each time slice. This combination of appearance and motion representation helps achieve fine-grained temporal alignment. Second, to choose among the appearance or motion tube in the multi-hop reasoning process, a switch module is implemented. Extensive experiments for the proposed method [17] have been carried out on the two benchmark datasets - MSVD-QA [7] and MSRVTT-QA [7].%The proposed model outperforms the previous state-of-the-art method GRAAM [7] with an improvement of overall accuracy by ~5\% on MSVD-QA dataset [7] and ~3\% on the MSRVTT-QA dataset [7].
\subsubsection{PAMN (Kim et al., 2019)}
[20] proposes the Progressive Attention Memory Network (PAMN) for movie story QA. PAMN [20] is designed to overcome 2 main challenges of movie story QA: finding out temporal parts for discovering relevant answers in case of movies that are longer in length and to fulfill the need to infer the answers of different questions on the basis of different modalities. PAMN [20] includes 3 main features: progressive attention mechanism which uses clues from video and question to find the relevant answer, dynamic modality fusion and belief correction answering scheme that improves the prediction on every answer. PAMN [20] performs experiments on two benchmark datasets: MovieQA [36] and TVQA [40]. %The experiments show that PAMN achieves state-of-the-art performance on MovieQA [36] and TVQA [40] datasets. [20] improves on [12] with an increase of 1.12\% in accuracy. 
\subsubsection{CAN (Yu et al., 2019)}
Earlier works in the domain of VQA have achieved favorable performance on short-term videos. Those approaches fail to work equally well on the long-term videos. Considering this drawback, [22] proposes a two-stream fusion strategy called compositional attention networks (CAN) for the VQA tasks. The method comprises of an encoder and a decoder. The encoder uses a uniform sampling stream (USS) which extracts the global semantics of the entire video and the action pooling stream (APS) is responsible for capturing the long-term temporal dynamics of the video. The decoder utilizes a compositional attention module (CAN) to combine the two-stream features generated by encoder using attention mechanism. The model is evaluated on two-short term (MSRVTT-QA [7] and MSVD-QA [7]) and one long-term (ActivityNet-QA [35]) VQA datasets. %The results indicate that CAN achieves a higher overall accuracy of 0.7 point on MSRVTT-QA [7] and 0.44 point on MSVD-QA [7] datasets.
\subsection{Multimodal Attention based Methods}
The task of VQA demands representations from multiple modalities such as videos and texts. Therefore, several approaches have worked upon the concept of multimodal attention to fuse the extracted knowledge from both modalities and obtain accurate answers. 
\subsubsection{MDAM (Kim et al., 2019)}
A video story QA architecture is proposed by [12], in which authors improve on [8] by proposing a Multimodal Dual Attention Model (MDAM). The main idea behind [12] was to incorporate a dual attention method with late fusion. The proposed model uses self attention mechanism to learn the latent concepts from video frames and captions. For a question, the model uses the attention over latent concepts. After performing late fusion, multimodal fusion is done. The main objective of [12] was to avoid the risk of over-fitting by using multimodal fusion methods such as concatenation or multimodal bilinear pooling. Using this technique, MDAM [12] learns to infer joint level representations from an abstraction of the video content. Experiments are conducted on the PororoQA and MovieQA datasets. 
\subsubsection{Chao et al., 2019}
One of the main challenges in VQA is that for a long video, the question may correspond to only a small segment of the video. Thus, it becomes inefficent to encode the whole video using an RNN. Therefore, authors of [14], propose a Question-Guided Video Representation module which summarizes the video frame-level features by employing an attention mechanism. Multimodal representations are then generated by fusing video summary and question information. The multimodal representations and dialogue-contexts are then passed with attention as an input in seq2seq model for answer generation. The authors conduct experiments using Audio-Visual Scene-Aware Dialog dataset (AVSD) [46] for evaluating their approach. %The model outperforms existing approaches on all evaluation metrics. 
\subsubsection{AMN (Yuan et al., 2020)}
In [23], authors present a framework Adversarial Multimodal Network (AMN) to better understand movie story QA. Most movie story QA methods proposed by [8],[12],[20] have 2 limitations in common. The models fail to learn coherent representations for multimodal videos and corresponding texts. In addition to that, they neglect the information that retains story cues. AMN [23] learns multimodal features by finding a coherent subspace for videos and the texts corresponding to those videos based on Generative Adversarial Networks (GANs). The model consists of a generator and a discriminator which constitute a couple of adversarial modules. The authors compare AMN [23] with other extended Image-QA baseline models and previous state-of-the-art methods for MovieQA [36] and TVQA [40] datasets. %AMN surpasses PAMN [20] with an overall increase of 2.78\% in accuracy.
\subsubsection{VQA-HMAL (Zhao et al., 2020)}
[24] proposes a method for open-ended long-form VQA which works upon the limitations of the short-form VQA methods. Short-form VQA methods do not consider the semantic representation in the long-form video contents hence they are insufficient for modeling long-form videos. The authors of [24] implement a hierarchical attentional encoder network using adaptive video segmentation which jointly extracts the video representations with long-term dependencies based on the questions. In order to generate the natural language answer, a decoder network based on a Generative Adversarial Network (GAN) is used on the multimodal text-generation task. Another major contribution of this paper accounts for generating three large-scale datasets for VQA from the ActivityNet [50], MSVD [49], TGIF [51] datasets.
\subsubsection{Jin et al., 2019}
[19] proposes a novel multi-interaction network that uses two different types of interactions, i.e. multimodal and multi-level interaction. The interaction between the visual and textual information is termed as multimodal interaction. Multi-level interaction occurs inside the multimodal interaction. The proposed new attention mechanism can simultaneously capture the element-wise and the segment-wise sequence interactions. Moreover, fine-grained information is extracted using object-level interactions that aid in fetching object motions and their interactions. The model is evaluated over the TGIF-QA, MSVD-QA and MSRVTT-QA datasets. 
\subsection{Miscellaneous Models}
In addition to the spatio-temporal, memory-based, attention-based and multimodal attention based models, the following methods use different novel approaches to solve the problem of VQA.

\subsubsection{MAR-VQA(a+m) (Zhuang et al., 2020)}
[25] takes into consideration multiple channels of information present in a video such as audio, motion and appearance. It differs from earlier works which either represented the question as a single semantic vector or directly fused the features of appearance and motion leading to loss of important information. [25] also incorporates external knowledge from Wikipedia which suggests attribute text related to the videos and aids the proposed model with support information. The model is experimented on two datasets - TGIF-QA and ActivityNet-QA. 
\subsubsection{MQL (Lei et al., 2020)}
A majority of previous works in VQA consider video-question pairs separately during training. However,there could be many questions addressed to a particular video and most of them have abundant semantic relations. In order to explore these semantic relations, [26] proposes Multi-Question Learning (MQL) that learns multiple questions jointly with their candidate answers for a particular video sequence. These joint learned representations of video and question can then be used to learn new questions. [26] introduces an efficient framework and training paradigm for MQL, where the relations between video and questions are modeled using attention network. This framework enables the co-training of multiple video-question pairs. MQL [26] is capable to perform better understanding and reasoning even in the case of a single question. Authors carry out extensive experiments on 2 popular datasets- TVQA and CPT-QA.
\subsubsection{Yang et al., 2020}
In natural language processing, BERT has outperformed the LSTM in several tasks. However, [27] presents one of the first works which explores BERT in the domain of computer vision. [27] proposes to use BERT representations to capture both the language and the visual information with the help of a pre-trained language-based transformer. Firstly, Faster-RCNN is used to extract the visual semantic information from video frames as visual concepts. Next, two independent flows are developed to process the visual concepts and the subtitles along with the questions and answers. In the individual flows, a fine-tuned BERT network is present to predict the right answer. The final prediction is obtained by jointly processing the output of the two independent flows. The proposed model is evaluated on TVQA [40] and Pororo [8] datasets.
\subsubsection{HCRN (Le et al., 2020)}
[29] proposes a general-purpose reusable Conditional Relation Network (CRN) that acts as a foundational block for constructing more complex structures for reasoning and representation on video. CRN [29] takes an input array of tensorial objects with a conditioning feature and computes an output array of encoded objects. The model consists of a stack of these reusable units for capturing contextual information and diverse modalities. This structural design of the model bolsters multi-step and high-order relational reasoning. CRN [29] hierarchy constitutes 3 main layers for capturing clip motion as context, linguistic context and video motion as context. One important advantage of CRN over [10],[16] and [13] is that it scales well on long-sized video simply with the addition of more layers in the structure. Performance evaluation methods demonstrate that CRN [29] achieves high performance accuracy as compared to various state-of-the-art models. CRN [29] is evaluated on 3 benchmark datasets- TGIF-QA [44], MSRVTT-QA [7] and MSVD-QA [7].

\section{Discussion}
Video Question Answering (VQA) is a complex and challenging task in the domain of Artificial Intelligence. Extensive literature is available for VQA and to the best of our knowledge no previous survey has been conducted for the same. In this survey, we studied the prominent datasets and models that have been researched in VQA. We also compare the results of all the models in tables 1, 2, 3 and 4 by grouping them according to the datasets. Significant improvements can be witnessed in the past years in the field of VQA. Therefore, there lies a huge potential for research in this field for future tasks.

\end{document}